\pdfoutput=1

\documentclass[11pt]{article}

\usepackage[]{naacl2021}

\usepackage{times}
\usepackage{latexsym}

\usepackage[T1]{fontenc}

\usepackage[utf8]{inputenc}

\usepackage{microtype}

\usepackage{subcaption}
\usepackage{graphicx}
\usepackage{amsmath}
\usepackage{nccmath}
\usepackage{amsfonts}
\usepackage{hhline}

\newcommand\numberthis{\addtocounter{equation}{1}\tag{\theequation}}

\usepackage{enumitem}
\usepackage{mathrsfs}

\usepackage{booktabs}
\usepackage[belowskip=-3pt]{caption}
\usepackage{float}
\usepackage{titlesec}
\usepackage{capt-of}

\usepackage[most]{tcolorbox}
\usepackage[framemethod=TikZ]{mdframed}

\usepackage{array}
\usepackage{arydshln}
\title{Modeling Human Mental States with an Entity-based Narrative Graph}

\author{I-Ta Lee, Maria Leonor Pacheco, Dan Goldwasser\\
       Department of Computer Science\\
       Purdue University\\
       West Lafayette, IN, USA\\
       \{lee2226, pachecog, dgoldwas\}@purdue.edu
       }

\begin{document}
\maketitle
\begin{abstract}
Understanding narrative text requires capturing characters' motivations, goals, and mental states.
This paper proposes an Entity-based Narrative Graph (ENG) to model the internal-states of characters in a story. We explicitly model entities, their interactions and the context in which they appear, and learn rich representations for them. We experiment with different task-adaptive pre-training objectives, in-domain training, and symbolic inference to capture dependencies between different decisions in the output space. We evaluate our model on two narrative understanding tasks: predicting character mental states, and desire fulfillment, and conduct a qualitative analysis. 
\end{abstract}

\section{Introduction}

Understanding narrative text requires modeling the motivations, goals and internal states of the characters described in it. These elements can help explain intentional behavior and capture causal connections between the characters’ actions and their goals.  
While this is straightforward for humans, machine readers often struggle as a correct analysis relies on making long range common-sense inferences over the narrative text. Providing the appropriate narrative representation for making such inferences is therefore a key component. In this paper, we suggest a novel narrative representation model and evaluate it on two narrative understanding tasks, analyzing the characters' mental states and motivations~\cite{abdul2017emonet,rashkin-etal-2018-modeling,chen2020you}, and desire fulfillment~\cite{chaturvedi2016ask,rahimtoroghi-etal-2017-modelling}. 

\begin{figure}[ht]
\centering
\includegraphics[width=0.4\textwidth]{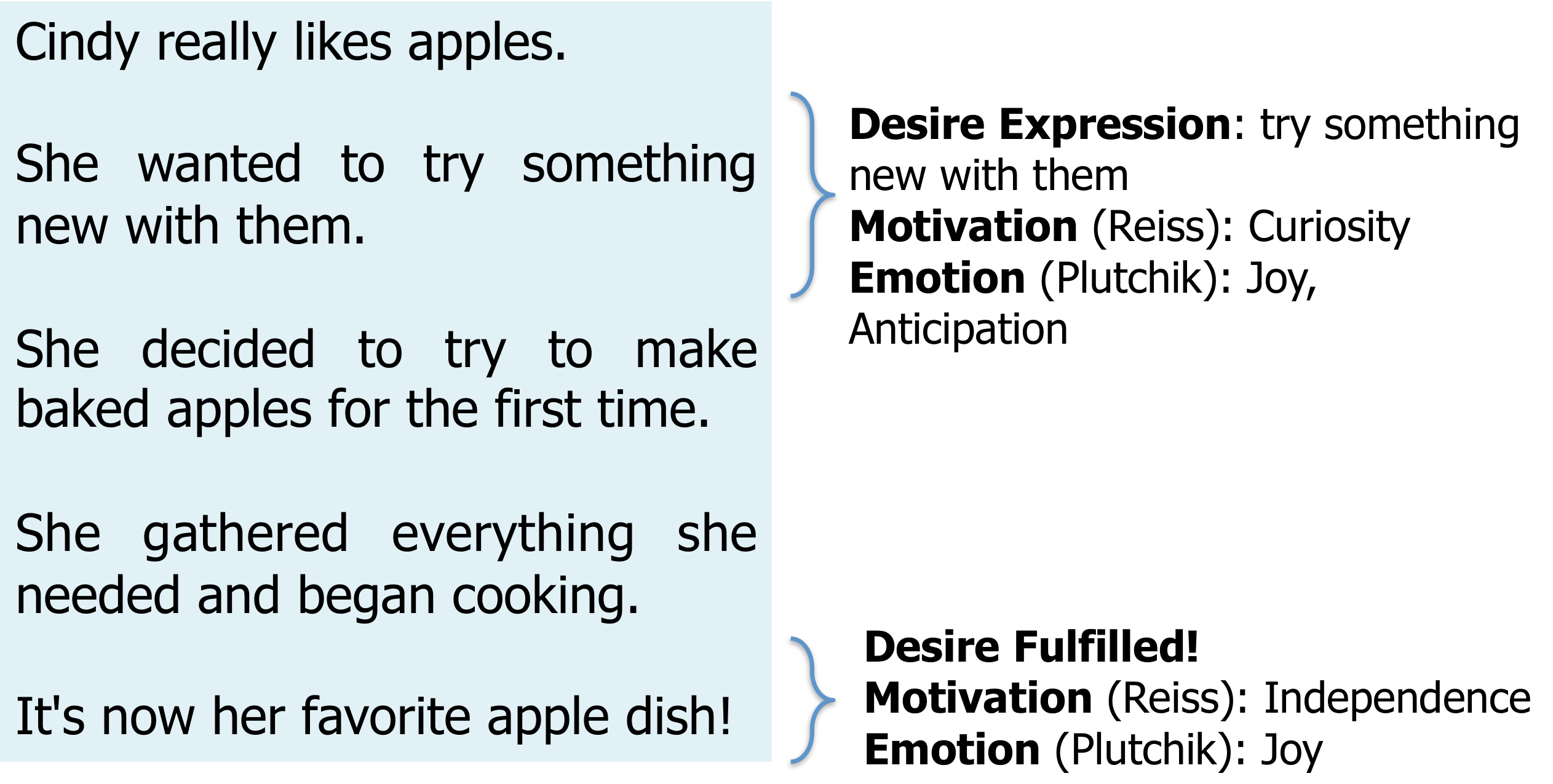}
\caption{Narrative Example}
\label{fig:narrativeExample}
\end{figure}

We follow the observation that narrative understanding requires an expressive representation capturing the context in which events appear and the interactions between characters' states. To clarify, consider the short story in Fig.~\ref{fig:narrativeExample}. The desire expression appears early in the story and provides the context explaining the protagonist's actions. 
Evaluating the fulfilment status of this expression, which tends to appear towards the end of the story, requires models that can reason over the desire expression (\textit{``trying something new''}), its target (\textit{``apples''}) and the outcome of the protagonist's actions (\textit{``it's now her favorite apple dish!''}). 
Capturing the interaction between the \textit{motivation} underlying the desire expression (in Fig.~\ref{fig:narrativeExample}, \textsc{Curiosity}) and the \textit{emotions} (in Fig.~\ref{fig:narrativeExample}, \textsc{Anticipation}) likely to be invoked by the motivation can help ensure the consistency of this analysis and improve its quality.


To meet this challenge, we suggest a graph-contextualized representation for entity states. Similar to contextualized word representations~\cite{peters2018deep,devlin2018bert}, we suggest learning an entity-based representation which captures the narrative it is a part of. For example, in \textit{``She decided to try to make baked apples for the first time”} the mental state of ``she'' would be represented differently given a different context, such as a different motivation for the action (\textit{``Her mother asked her to make an apple dish for a dinner party''}). In this case, the contextualized representation would capture the different emotion associated with it (e.g., \textsc{Fear} of disappointing her mother).
Unlike contextualized word embeddings, \textit{entity-based} contextualization needs to consider, at least, two levels of context: local text context and distant event context, which require more complicated modeling techniques to capture event semantics.
Moreover, the context of event relationships can spread over a long narrative, exceeding maximum sequence length limitation in modern contextualized word embedding models such as BERT~\cite{devlin2018bert}.

In this paper, we propose an Entity-based Narrative Graph (ENG) representation of the text. Unlike other graph-based narrative representations~\cite{lehnert1981plot,goyal2010automatically,elson2012dramabank} which require intensive human annotation, we design our models around low-cost supervision sources and shift the focus from symbolic graph representations of nuanced information to their learned embedding. 
In ENG, each node is associated with an entity-event pair, representing an entity mention that is involved in an event. Edges represent observed relations between entities or events.
We adapt the definition of event relationships introduced in \citet{lee2020ng} to our entity-event scenario. For entity relationships, the \textbf{CNext} relationship connects two coreferent entity nodes. For event relationships, the \textbf{Next} relationship captures the sequential order of events as they appear in the text, and six discourse relation types from the Penn Discourse Tree Bank (PDTB)~\cite{prasad2007penn} are used. These include \textbf{Before, After, Sync., Contrast, Reason} and \textbf{Result}. Note that these are extracted in a weakly supervised manner, without expensive human annotations.


%

To contextualize the entity embeddings over ENG, we apply a Relational Graph Convolution Network (R-GCN)~\cite{schlichtkrull2018modeling}, a relational variant of the Graph Convolution Network architecture (GCN)~\cite{kipf2016semi}. R-GCNs create contextualized node representations by considering the graph structure through graph convolutions and learn a composition function. This architecture allows us to take into account the narrative structure and the different discourse relations connecting the entity-event nodes. 


To further enhance our model, we investigate three possible pre-training paradigms: whole-word-masking, node prediction, and link prediction. All of them are constructed by automatically extracting noisy supervision and pre-training on a large-scale corpus. We show that choosing the right pre-training strategy can lead to significant performance enhancements in downstream tasks. For example, automatically extracting sentiment for entities can impact downstream emotion predictions.
Finally, we explore the use of a symbolic inference layer to model relationships in the output space, and show that we can obtain additional gains in the downstream tasks that have strong correlation in the output space. 

The evaluated downstream tasks include two challenging narrative analysis tasks, predicting characters' psychological states~\cite{rashkin-etal-2018-modeling} and  desire fulfilment~\cite{rahimtoroghi-etal-2017-modelling}. Results show that our model can outperform competitive transformer-based representations of the narrative text, suggesting that explicitly modeling the relational structure of entities and events is beneficial. Our code and trained models are publicly available\footnote{\url{https://github.com/doug919/entity_based_narrative_graph}}.

\section{Related Work} \label{sec:related}

Tracking entities and modeling their properties has proven successful in a wide range of tasks, including language modeling \cite{ji-etal-2017-dynamic}, question answering \cite{DBLP:conf/iclr/HenaffWSBL17} and text generation \cite{DBLP:conf/iclr/BosselutLHEFC18}. In an effort to model complex story dynamics in text, \citet{rashkin-etal-2018-modeling} released a dataset for tracking the emotional reactions of characters in stories. In their dataset, each character mention is annotated with three types of mental state descriptors: Maslow's ``hierarchy of needs''~\cite{maslow:motivation}, Reiss' ``basic motives''~\cite{reiss2004}, that provide a more informative range of motivations, and Plutchik's ``wheel of emotions''~\cite{plutchik1980general}, comprised of eight basic
emotional dimensions (e.g. joy, sadness, etc). In their paper, they showed that neural models with explicit or latent entity representations achieve promising results on this task. \citet{paul-frank-2019-ranking} approached this task by extracting multi-hop relational paths from ConceptNet, while \citet{gaonkar-etal-2020-modeling} leveraged semantics of the emotional states by embedding their textual description and modeling the co-relation between different entity states. \citet{rahimtoroghi-etal-2017-modelling} introduced a dataset for the task of desire fulfillment. They identified desire expressions in first-person narratives and annotated their fulfillment status. They showed that models that capture the flow of the narrative perform well on this task. 

Representing the narrative flow of stories using graph structures and multi-relational embeddings has been studied in the context of script learning \cite{DBLP:conf/ijcai/LiDL18,lee2019multi,lee2020ng}. In these cases, the nodes represent predicate-centric events, and entity mentions are added as context to the events. In this paper, we use an entity-centric narrative graph, where nodes are defined by entity mentions and their textual context. We encode the textual information in the nodes using pre-trained language models~\cite{devlin2018bert,liu2019roberta}, and the graph structure with a relational graph neural network~\cite{schlichtkrull2018modeling}. To learn the representation, we incorporate a task-adaptive pre-training phase. \citet{gururangan-etal-2020-dont} showed that further specializing large pre-trained language models to domains and tasks within those domains is effective.  





\section{Entity-based Narrative Graph}

\subsection{Framework Overview}

%
Many NLU applications require understanding entity states in order to make sophisticated inferences~\cite{sap2018atomic,bosselut2019comet,rashkin-etal-2018-modeling}, and the entity states are highly related to the event the entity involves in. 
%
In this work, we propose a learning framework that aims at modeling entities' internal states, and their interactions to other entities' internal states through events.
We include task-adaptive pre-training (TAPT) and downstream task training to train an entity-based narrative graph (ENG), a graph neural model designed to capture implicit states and interactions between entities. We extend the narrative graph proposed by \citet{lee2020ng}, which models event relationships, and instead of learning node representations for events, we focus on entity mentions that are involved in events. This change is motivated by the high-demand of NLU applications that require understanding entity mentions' states in order to make sophisticated inference.
%
%

\begin{figure}[h]
\centering
\includegraphics[width=0.32\textwidth]{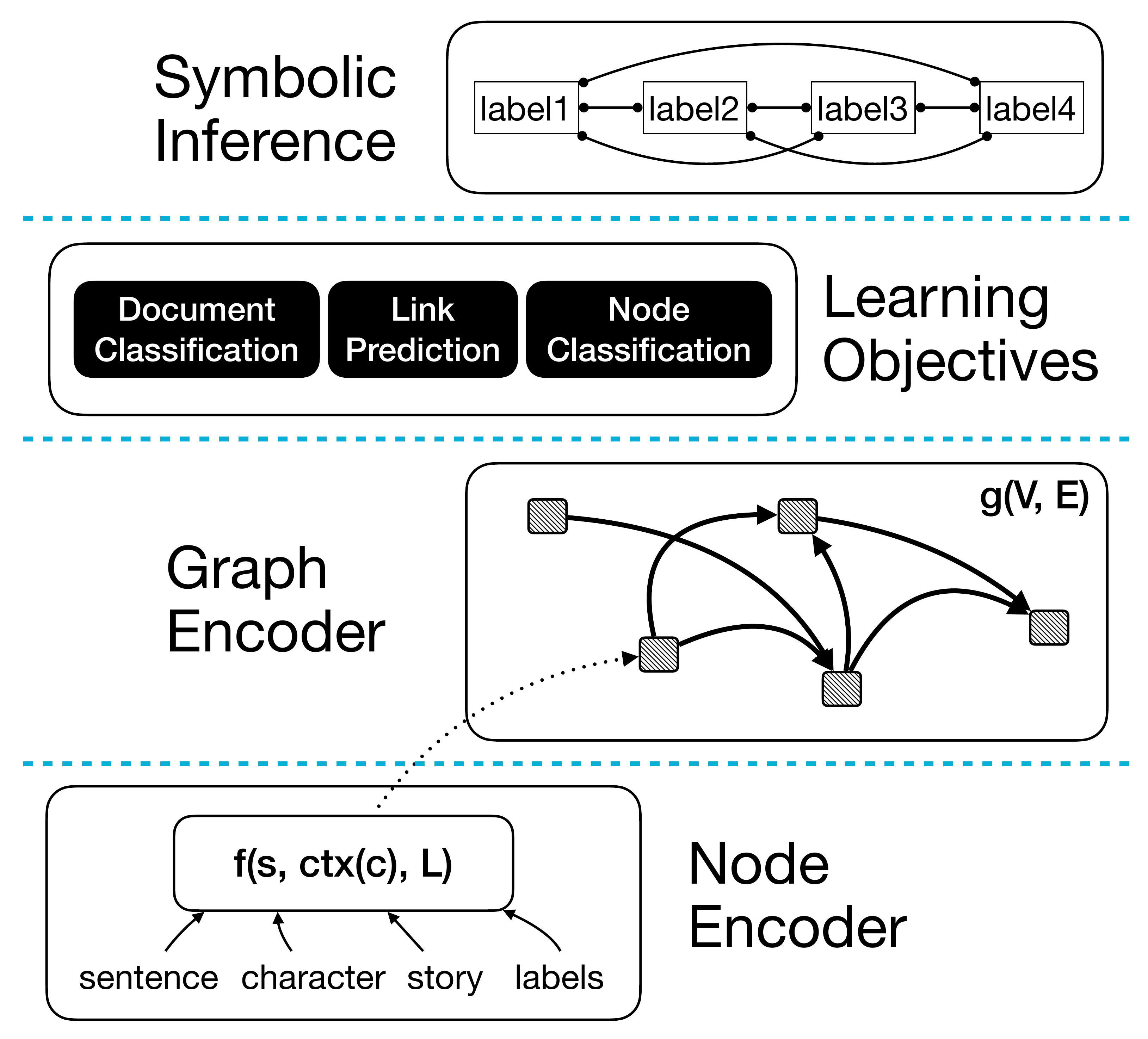}
\caption{Overview of the ENG framework.}
\label{fig:overview}
\end{figure}

Our framework consists of four main components: Node Encoder, Graph Encoder, Learning Objectives, and Symbolic Inference, outlined in Figure \ref{fig:overview}. The node encoder is a function used to extract event information about the target entity mention corresponding to the local node representation. The graph encoder uses a graph neural network to contextualize node representations with entity-events in the same document, generating entity-context-aware representations. The learning objectives use this representation for several learning tasks, such as node classification, link prediction, and document classification. Finally, we include a symbolic inference procedure to capture dependencies between output decisions.

We introduce a training pipeline, containing pre-training and downstream training, following recent evidence suggesting that task-adaptive pre-training is potentially useful for many NLU tasks~\cite{gururangan-etal-2020-dont}. We experiment with three pre-training setups, including the common whole-word-masking pre-training~\cite{liu2019roberta}, and two newly proposed unsupervised pre-training objectives based on ENG. We then evaluate two downstream tasks: StoryCommonsense~\cite{rashkin-etal-2018-modeling} and DesireDB~\cite{rahimtoroghi-etal-2017-modelling}. StoryCommonsense aims at predicting three sets of mental states based on psychological theories~\cite{maslow:motivation,reiss2004,plutchik1980general}, while DesireDB's goal is to identify whether a target desire is satisfied or not. Solving these tasks requires understanding entities' mental states and their interactions.

\subsection{Node Encoder}

Each node in our graph captures the local context of a specific entity mention (or character mention), and how the entity mentions are extracted is related to extracting their edges, which will be described in Sec. \ref{sec:graph_encoder}.
Following \citet{gaonkar-etal-2020-modeling}, we format the input information to be fed into a pretrained language model. For a given character $c$ and sentence $s$, the inputs to the node encoder consist of three components $(s, ctx(c), L)$, where $s$ is the sentence in which $c$ appears, $ctx(c)$ is the context of $c$ (all the sentences that the character appears in), and $L$ is a label sentence. The label sentence is an artificial sentence of the form ``[entity name] is [label 1], [label 2], ..., [label k].'' The $k$ labels correspond to the target labels in the downstream task. For example, in StoryCommonsense, the Plutchik state prediction task has eight labels characterizing human emotions, such as \textit{joy}, \textit{trust}, and \textit{anger}. \citet{gaonkar-etal-2020-modeling} show that self-attention is an effective way to let the model take label semantics into account, and improve performance\footnote{Note that all candidate labels are appended to every example, without denoting which one is the right answer. Our preliminary experiments confirm that taking label semantics into account improves performance}.

Our best model uses RoBERTa~\cite{liu2019roberta}, a highly-optimized version of BERT~\cite{devlin2018bert}, to encode nodes. 
We convert the node input $(s, ctx(c), L)$ to RoBERTa's two-sentence input format by treating $s$ as the first sentence, and the concatenation of $ctx(c)$ and $L$ as the second sentence. After forward propagation, we take the pooled sentence representation (i.e., \textless \textit{s} \textgreater for RoBERTa, \textit{CLS} for BERT), as the node representation $v$. This is formulated as $v = f_{roberta}(s, ctx(c), L)$.

\subsection{Graph Encoder} \label{sec:graph_encoder}

The ENG is defined as $ENG=(V, E)$, where $V$ is the set of encoded nodes in a document and $E$ is the set of edges capturing relationships between nodes. Each edge $e \in E$ is a triplet $(v_1, r, v_2)$, where $v_1, v_2 \in V$ and $r$ is an edge type ($r \in R$). Following \citet{lee2020ng}, we use eight relation types ($|R| = 8$) that have been shown to be useful for modeling narratives. \textsc{Next} denotes if two nodes appear in neighboring sentences. \textsc{CNext} expresses the next occurrence of a specific entity following its co-reference chain. 
Six discourse relation types, used by \citet{lee2020ng} and defined in Penn Discourse Tree Bank (PDTB)~\cite{prasad2007penn}, are also used in this work, including \textsc{Before}, \textsc{After}, \textsc{Sync.}, \textsc{Contrast}, \textsc{Reason}, \textsc{Result}. Their corresponding definition in PDTB and can be found in Table \ref{tab:disc_relations}.
Following \citet{lee2020ng}, we use the Stanford CoreNLP pipeline\footnote{Stanford CoreNLP v4.0 with default annotators.} ~\cite{manning2014stanford} to obtain co-reference links and dependency trees. We use them as heuristics to extract the above relations and identify entities for TAPT\footnote{For StoryCommonsense, since the entity names are annotated, we simply use them.}. 
Details of this procedure can be found in \cite{lee2020ng}. Note that although we share the same relation definitions, our nodes are defined over entities, instead of events.

\begin{table}[h]\small
   \centering 
   \small
   \begin{tabular}{lll} 
   \toprule
   \textbf{Abbrev.} & \textbf{PDTB} & \textbf{Distr.}\\
   \midrule
    \textsc{Next} & -- & 50\% \\
    \textsc{CNext} & -- & 20\% \\
    \textsc{Before} & Temporal.Async.Precedence & 5\% \\
    \textsc{After} & Temporal.Async.Succession & 5\% \\
    \textsc{Sync.} & Temporal.Synchrony & 5\% \\ 
    \textsc{Contrast}  & Comparison.Contrast & 5\% \\ 
    \textsc{Reason}  & Contingency.Cause.Reason & 5\% \\
    \textsc{Result} & Contingency.Cause.Result & 5\% \\ 
   \bottomrule
   \end{tabular}
   \caption{Alignment between PDTB relations and the abbreviations used in this paper. The third column in the sampling distribution.}
   \label{tab:disc_relations}
\end{table}

For encoding the graph, we use a Relational Graph Convolution Network (R-GCN)~\cite{schlichtkrull2018modeling}, which is designed for Knowledge Base Completion. This architecture is capable of modeling typed edges and is resilient to noise. R-GCN is defined as:
\vspace{-5pt}
\begin{align*}\small
h^{l+1}_i &= ReLU\Bigg( \sum_{r \in R} \sum_{u \in U_r(v_i)} \frac{1}{z_{i,r}} W_{r}^{l} h_{u}^{l} \Bigg),  \numberthis \label{eq:R-GCN} 
\end{align*}
where $h_i^{l}$ is the hidden representation for the i-th node at layer $l$ and $h_i^0 = v_i$ (output of the node encoder); $U_r(v_i)$ represents $v_i$'s neighboring nodes connected by the relation type $r$; $z_{i, r}$ is for normalization; and $W_r^l$ represents trainable parameters.

Our implementation of R-GCN propagates messages between entity nodes, emulating the interactions between their psychological states, and thus enriching node representations with context. Note that our framework is flexible, and alternative node and graph encoders could be used. 

\subsection{Output Layers and Learning Objectives} \label{sec:objectives}

We explore three learning problem types. 

\paragraph{Node Classification} 
For node classification, we use the contextualized node embeddings coming from the graph encoder, and plug in a $k$-layer feed-forward neural network on top ($k=2$ in our case). The learning objectives could be either multi-class or multi-label. For multi-class classification, we use the weighted cross-entropy loss (CE). For multi-label classification, we use the binary cross-entropy (BCE) loss for each label\footnote{We tried weighted an unweighted BCE, and selected the unweighted one for our final model.}:
\vspace{-5px}
\begin{align*}
CE &= - \frac{1}{N} \sum_{i=1}^N \alpha_i y_i \log(S(g(f(x_i)))),  \numberthis \label{eq:ce_loss} 
\end{align*}
where $S(.)$ is the Softmax function, $f(.)$ is the graph encoder, $g(.)$ is the node encoder, $x_i$ is the input including the target node $i$ ($(s, ctx(c), L)$) and all other nodes in the same document (or ENG), $y_i$ is the label, and $\alpha_i$ is the example weight based on the label distribution of the training set..

\paragraph{Link Prediction}
This objective tries to recover missing links in a given ENG. We sample a small portion of edges ($20\%$ in our case) as positive examples, based on the relation type distribution given in Table \ref{tab:disc_relations}, taken from the training set. 
To obtain negative examples, we corrupt the positive examples by replacing one component of the edge triplet with a sampled component so that the resulting triplet does not exist in the original graph. For example, given a positive edge $(e_1, r, e_2)$, we can create negative edges: $(e_1’, r, e_2)$, $(e_1, r’, e_2)$, or $(e_1, r, e_2’)$.
Following \citet{schlichtkrull2018modeling}, we score each edge sample with DistMult~\cite{chang2014typed}:
\vspace{-5pt}
\begin{align*}
D(i, r, j) &= h_i^T W_r h_j, \numberthis \label{eq:distmult} 
\end{align*}
where $W_r$ is a relation-specific trainable matrix (non-diagonal) and $h_i$ and $h_j$ are node embeddings coming from the graph encoder. A higher score indicates that the edge is more likely to be active. To learn this, we reward positive samples and penalize negative ones, using an adapted CE loss:
\vspace{-5px}
\begin{align*}
  L = - \frac{1}{|T|} \sum_{(i, r, j, y) \in T} y \log( \sigma(\epsilon_r D(i, r, j)) ) \\
  + (1-y) \log( 1 - \sigma(\epsilon_r D(i, r, j)) ), \numberthis \label{eq:loss}
\end{align*}
$T$ is the sampled edges set, $y = \{0, 1\}$,  $\sigma(.)$ is the Sigmoid function, and $\epsilon_r$ is the edge type weight, based on the edge sampling rate in Table \ref{tab:disc_relations}.

\paragraph{Document Classification} 
For document classifications, such as DesireDB, we aggregate the node representations from the entire ENG to form a single representation. To leverage the relative importance of each node, we add a self-attention layer on top of the graph nodes. We calculate the attention weights by attending on the query embedding (in DesireDB, this is the sentence embedding for the desire expression). 
\vspace{-5px}
\begin{align*}
  a_i &= ReLU(W_a [h_i; h_t] + b_a) \\
  z_i &= exp(a_i) \\
  \alpha_i &= \frac{z_i}{\sum_{k} z_k} \;\;\;;\;\;\;\; h_d = \sum_i \alpha_i h_i \numberthis \label{eq:attn}
\end{align*}
where $h_i$ is the i-th node representation, $h_t$ is the query embedding, $W_a$ and $b_a$ are trainable parameters, and $h_d$ is the final document representation. We then feed $h_d$ to a two-hidden-layer classifier to make predictions. We use the loss function specified in Eq. \ref{eq:ce_loss}.

\subsection{Task-Adaptive Pre-training}\label{sec:pretraining}
Recent studies demonstrate that downstream tasks performance can be improved by performing self-supervised pre-training on the text of the target domain~\cite{gururangan-etal-2020-dont}, called Task-Adaptive Pre-Training (TAPT).
To investigate whether different TAPT objectives can provide different insights for downstream tasks, we apply three possible pre-training paradigms and compare them on StoryCommonsense. We focus on StoryCommonsense given that the dataset was created by annotating characters' mental states on a subset of RocStories~\cite{mostafazadeh2016corpus}, a corpus with 90K short common-sense stories. This provides us with a large unlabeled resource for investigating different pre-training methods. We run TAPT on all the RocStories text\footnote{Not including the validation and testing sets of Story Cloze Test}. We use the learning parameters suggested by ~\citet{gururangan-etal-2020-dont} and explore the following strategies:

\textbf{Whole-Word Masking:} Randomly masks a subset of words and asks the model to recover them from their context~\cite{radford2019language,liu2019roberta}. We perform this task over RoBERTa, initialized with \textit{roberta-base}.

\textbf{ENG Link Prediction:} Weakly-supervised TAPT over the ENG. The setup follows Sec. \ref{sec:objectives} (Link Prediction) to learn a model that can recover missing edges in the ENG. 

\textbf{ENG Node Sentiment Classification:} Performs weakly-supervised sentiment TAPT. We use the Vader sentiment analysis~\cite{hutto2014vader} tool to annotate the sentiment polarity for each node in the ENG, based on its sentence. The setup follows Sec. \ref{sec:objectives} (Node Classification). 

\subsection{Symbolic Inference}\label{sec:symbolic}

In addition to modeling the narrative structure in the embedding space, we add a symbolic inference procedure to capture structural dependencies in the output space for the StoryCommonsense task. To model these dependencies, we use DRaiL \cite{PG_tacl_20}, a neural-symbolic framework that allows us to define probabilistic logical rules on top of neural network potentials. 

Decisions in DRaiL are modeled using rules, which can be weighted (i.e., soft constraints), or unweighted (i.e., hard constraints). Rules are formatted as horn clauses: \textsc{A} $\Rightarrow$ \textsc{B}, where \textsc{A} is a conjunction of observations and predicted values, and \textsc{B}  is the output to be predicted. Each weighted rule is associated with a neural architecture, which is used as a scoring function to obtain the rule weight. The collection of rules represents the global decision, and the solution is obtained by
performing MAP inference. Given that rules are written as horn clauses, they can be expressed as linear inequalities corresponding to their disjunctive form, and thus MAP inference is defined as a linear program. 

In DRaiL, parameters are trained using the structured hinge loss. This way, all neural parameters are updated to optimize the global objective. Additional details can be found in \cite{PG_tacl_20}. To score weighted rules, we used feed-forward networks over the node embeddings obtained by the objectives outlined in Sec. \ref{sec:objectives} and \ref{sec:pretraining}, without back-propagating to the full graph. We model the following rules:


\paragraph{Weighted rules} We score each state, as well as \textit{state transitions} to capture the progression in a character's mental state throughout the story. 

\begin{equation*}\small
\begin{split}
& \mathtt{Entity(e_i)} \Rightarrow \mathtt{State(e_i,l_i)} \\
& \mathtt{State(e_i,l_i)} \wedge \mathtt{HasNext(e_i,e_j)} \Rightarrow \mathtt{State(e_j,l_j)} \\
\end{split}
\end{equation*}

\noindent where $e_i$ and $e_j$ are two different mentions of the same character, and $\mathtt{HasNext}$ is a relation between consecutive sentences. $\mathtt{State}$ can be either $\mathtt{Maslow}$, $\mathtt{Reiss}$ or $\mathtt{Plutchik}$.

\paragraph{Unweighted rules} There is a dependency between Maslow’s ``hierarchy of needs' and Reiss ``basic motives''~\cite{rashkin-etal-2018-modeling}. We introduce logical constraints to disallow mismatches in the Maslow and Reiss prediction for a given mention $e_i$. In addition to this, we model positive and negative sentiment correlations between Plutchik labels. To do this, we group labels into positive (e.g. joy, trust), and negative (e.g. fear, sadness). We refer to this set of rules as \textit{inter-label dependencies}.
\begin{equation*}\small
\begin{split}
& \mathtt{Maslow(e_i,m_i)} \wedge \neg\mathtt{Align(m_i,r_i)} \Rightarrow \neg\mathtt{Reiss(e_i,r_i)} \\
& \mathtt{Reiss(e_i,r_i)} \wedge \neg\mathtt{Align(m_i,r_i)} \Rightarrow \neg \mathtt{Maslow(e_i,m_i)} \\
& \mathtt{Plut(e_i,p_i)} \wedge \mathtt{Pos(p_i)} \wedge \neg\mathtt{Pos(p_j)} \Rightarrow \neg\mathtt{Plut(e_i,p_j)}
\end{split}
\end{equation*}
%


Given that the DesireDB task requires a single prediction for each narrative graph, we do not employ symbolic inference for this task.

\section{Evaluation} \label{sec:eval}

\begin{table*}[ht]\centering
\small
\setlength\tabcolsep{5pt}
\begin{tabular}{@{}crrrrrrrrrrr@{}}\toprule
&& \multicolumn{3}{c}{\textbf{Maslow}} & \multicolumn{3}{c}{\textbf{Reiss}} & \multicolumn{3}{c}{\textbf{Plutchik}} \\ 
\cmidrule{3-12}
\textbf{Group} & \textbf{Models} & \textbf{Precision} & \textbf{Recall} & \textbf{F1} & \textbf{Precision} & \textbf{Recall} & \textbf{F1} & \textbf{Precision} & \textbf{Recall} & \textbf{F1} \\ \midrule 
G1 &\textbf{\textsc{Random}} &  7.45 & 49.99  & 12.96 & 1.76 & 50.02  & 3.40 & 10.35 & 50.00  & 17.15 \\
&\textbf{\textsc{TF-IDF}} &  29.79 & 34.56  & 32.00 & 20.55  & 24.81 & 22.48 & 22.71  & 25.24 & 23.91 \\
&\textbf{\textsc{GloVe}} &  27.02 & 37.00  & 31.23 & 16.99  & 26.08 & 20.58 & 19.47  & 46.65 & 27.48 \\
&\textbf{\textsc{LSTM}} &  30.34 & 40.12  & 34.55 & 21.38  & 28.70 & 24.51 & 25.31  & 33.44 & 28.81 \\
&\textbf{\textsc{CNN}} &  29.30 & 44.18  & 35.23 & 17.87  & 37.52 & 24.21 & 24.47  & 38.87 & 30.04 \\
&\textbf{\textsc{REN}} &  26.85  & 44.78 & 33.57 & 16.73 & 26.55 & 20.53 & 25.30 & 37.30 & 30.15 \\ 
&\textbf{\textsc{NPN}} &  26.60 & 39.17  & 31.69 & 15.75  & 20.34 & 17.75 & 24.33  & 40.10 & 30.29 \\\midrule
G2&\textbf{SA-ELMo*} &  34.91 & 32.16  & 33.48 & 21.23 & 16.53  & 18.59 & 47.33 & 40.86  & 43.86 \\
&\textbf{SA-RBERT*} & 43.58 & 30.03  & 35.55 & 24.75 & 18.00  & 20.84 & 46.51 & 45.45  & 45.97 \\
&\textbf{LC-BERT*} &  43.05 & 41.31  & 42.16 & 29.46 & 28.67  & 29.06 & 49.36 & 52.09  & 50.69 \\
&\textbf{LC-RBERT*} & 43.25 & 47.17  & 45.13 & 39.62 & 29.75  & 33.98 & 47.87 & 53.41  & 50.49 \\ \midrule
G3&\textbf{ENG} & 43.87 & 51.13  & 47.22 & 37.66 & 36.20  & 36.92 & 48.96 & 56.07  & 52.27 \\ \hdashline
&\textbf{ENG+Mask} & 44.27 & 53.54  & \textbf{48.47} & 39.29 & 33.93 & 36.41 & 49.64  & 56.93 & \textbf{53.03} \\
&\textbf{ENG+Link} & 43.47 & 52.80  & 47.68 & 37.17  & 37.18 & \textbf{37.18} & 50.62  & 54.48 & 52.48 \\
&\textbf{ENG+Sent} & 45.29 & 50.89  & 47.93 & 36.69  & 36.14 & 36.41 & 49.48  & 57.12 & \textbf{53.03} \\
\midrule
G4 &\textbf{ENG+IL} & 40.90 & 58.03 & 47.98 & 31.67 & 41.19 & 35.81 & 49.93 & 74.95 & 59.93   \\
&\textbf{ENG+IL+ST}  & 40.47 & 58.43 & 47.82 &  31.80 & 40.58 & 35.66 &   51.19 & 72.60 & \textbf{60.04} \\
\bottomrule
\end{tabular}
\caption{Results for the StoryCommonsense task, including three multi-label tasks (Maslow, Reiss, and Plutchik), for predicting human's mental states of motivations or emotions. The star sign indicates that the result is from our re-implemented version of previous baselines.}
\label{tab:state_result}
\end{table*}

Our evaluation includes two downstream tasks and a qualitative analysis. We report the results for different TAPT schemes and symbolic inference on StoryCommonsense. For the qualitative analysis, we visualize and compare the contextualized graph embeddings and contextualized word embeddings. 

\subsection{Data and Experiment Settings}

For TAPT, we use RocStories, as it has a decent amount of documents (90K after excluding the validation and testing sets) that share the text style of StoryCommonsense. For all tasks, we use the train/dev/test splits used in previous work.

All the RoBERTa models used in this paper are initialized with \textit{roberta-base}, and the BERT models with \textit{bert-base-uncased}. 
The maximum sequence length for the language models is $160$. If the input sequence exceeds this number, we will keep the label sentence untouched and cut down the main sentence.
For large ENGs, such as long narratives in DesireDB, we set the maximum number of nodes to $60$; all the hidden layer have $128$ hidden units; and the number of layers for R-GCN is $2$. 
For learning parameters in TAPT, we set the batch size to $256$ through gradient accumulations; the optimizer is Adam~\cite{kingma2014adam} with an initial learning rate of $\mathtt{1e-4}$, $\epsilon=\mathtt{1e-6}$, $\beta = (0.9, 0.98)$, weight decay $0.01$, and warm-up proportion $0.06$. We run TAPT for $100$ epochs. For the downstream tasks, we conduct a grid search of Adam's initial learning rate from $\{\mathtt{2e-3, 2e-4, 2e-5, 2e-6}\}$, $5000$ warm-up steps, and stop patience of $10$. Model selection is done on the validation set. We report results for the best model. For learning the potentials for symbolic inference with DRaiL~\cite{PG_tacl_20}, we use local normalization with a learning rate of $\mathtt{1e-3}$, and represent neural potentials using 2-layer Feed-Forward Networks over the ENG node embeddings. All hidden layers consist of 128 units. The parameters are learned using SGD with a patience of 5, tested against the validation set. For more details, refer to \cite{PG_tacl_20}. Note that while it would be possible to back-propagate  to the whole graph, this is a computationally expensive procedure. We leave this exploration for future work.


\subsection{Task: StoryCommonsense} \label{sec:storycommonsense}

StoryCommonsense consists of three subtasks: Maslow, Reiss, and Plutchik, introduced in Sec. \ref{sec:related}. 
Each subtask is a multi-label classification task, where the input is a sentence-character pair in a given story, and the output is a set of mental state labels.
Each story was annotated by three annotators and the final labels were determined through a majority vote. For Maslow and Reiss, the vote is count-based, i.e., if two out of three annotators flag a label, then it is an active label. For Plutchik, the vote is rating-based, where each label has an annotated rating, ranging from $\{0, 5\}$. If the averaged rating is larger or equal to $2$, then it is an active label. This is the set-up given in the original paper~\cite{rashkin-etal-2018-modeling}. Some papers~\cite{gaonkar-etal-2020-modeling} report results using only the count-based majority vote, resulting in scores that are not comparable to ours. Therefore, we re-implement two recent strong models proposed for this task. 
The Label Correlation model (LC~\cite{gaonkar-etal-2020-modeling}) applies label semantics as input and model output space using a learned correlation matrix.
The Self-Attention model (SA~\cite{paul-frank-2019-ranking}) utilize attentions over multi-hop knowledge paths extracted from external corpus. 
We evaluate them under the same set of hyper-parameters and model selection strategies as our models.

We briefly explain all the baselines, as well as our model variants shown in Table \ref{tab:state_result}. The first group (G1) are the baselines proposed in the task paper. \textbf{TF-IDF} uses TF-IDF features, trained on RocStories, to represent the target sentence $s$ and character context $ctx(c)$, and uses a Feed-Forward Net (FFN) classifier; \textbf{GloVe} encodes the sentences with the pretrained GloVe embeddings and uses a FFN; \textbf{CNN}~\cite{kim2014convolutional} replaces the FFN with a Convolutional Neural Network; \textbf{LSTM} is a two-layer bi-directional LSTM; \textbf{REN}~\cite{DBLP:conf/iclr/HenaffWSBL17} is a recurrent entity network that learns to encode information for memory cells; and \textbf{NPN}~\cite{DBLP:conf/iclr/BosselutLHEFC18} is an \textbf{REN} variant that includes a neural process network.

The second group (G2) of baselines are based on two recent publications--\textbf{LC} and \textbf{SA}--that showed strong performance on this task. We re-implement them and run the evaluation under the same setting as our proposed models. They originally use BERT and ELMo, respectively. To provide a fair comparison, we also train a RoBERTa variant for them (LC-RBERT and SA-RBERT). 
Note that the original paper of SA~\cite{paul-frank-2019-ranking} reports an F1 of $59.81$ on Maslow and $35.41$ on Reiss, while LC~\cite{gaonkar-etal-2020-modeling} reports $65.88$ on Plutchik. However, these results are not directly comparable to ours. The discrepancy arises mainly from two points: (1) The rating-based voting, described in Sec. \ref{sec:storycommonsense}, is not properly applied, and (2) We do not optimize the hyper-parameter search space in our setting, given the relatively expensive pre-training. Our re-implemented versions give a better foundation for a fair comparison.

The third (G3) and fourth (G4) groups are our model variants. \textbf{ENG} is the model without TAPT; \textbf{ENG+Mask}, \textbf{ENG+Link}, and \textbf{ENG+Sent} are the models with Whole-Word-Masking (WM), Link Prediction (LP), and Node Sentiment (NS) TAPT, respectively. In the last group, \textbf{ENG(Best) + IL} and \textbf{ENG(Best) + IL + ST} are based on our best ENG model with TAPT and adding inter-label dependencies (IL) and state transitions (ST) using symbolic inference, described in Sec. \ref{sec:symbolic}.

Table \ref{tab:state_result} reports all the results. We can see that Group 2 generally performs better than Group 1 on all three subtasks, suggesting that our implementation is reasonable. Even without TAPT, \textbf{ENG} outperforms all baselines, rendering $2-3\%$ absolute F1-score improvement. With TAPT, the performance is further strengthened. Moreover, we find that different TAPT tasks offer different levels of improvement for each subtask. The WM helps the most in Maslow and Plutchik, while the LP and NS excel in Reiss and Plutchik, respectively. This means that different TAPTs embed different information needed for solving the subtask. For example, the ability to add potential edges can be key to do motivation reasoning (Reiss), while identifying sentiment polarities (NS) can help in emotion analysis (Plutchik). This observation suggests a direction of connecting different related tasks in a joint pipeline. We leave this for future work.

Lastly, we evaluate the impact of symbolic inference. 
We perform joint inference over the rules defined in Sec. \ref{sec:symbolic}. On Table \ref{tab:state_result}, we can appreciate the advantage of modeling these dependencies for predicting Plutchik labels. However, the same is not true for the other two subtasks, where symbolic inference increases recall at the expense of precision, resulting in no F1 improvement. Note that labels for Maslow and Reiss are sparser, accounting     for 55\% and 42\% of the nodes, respectively. In contrast, Plutchik labels are present in 68\% of the nodes.

\subsection{Task: DesireDB}

\begin{table*}[]\centering
\small
\setlength\tabcolsep{5pt}
\begin{tabular}{@{}rrrrrrrrrr@{}}\toprule
& \multicolumn{3}{c}{\textbf{Fulfilled}} & \multicolumn{3}{c}{\textbf{Unfulfilled}} & \multicolumn{3}{c}{\textbf{Average}} \\ 
\cmidrule{2-10}
\textbf{Models} & \textbf{Precision} & \textbf{Recall} & \textbf{F1} & \textbf{Precision} & \textbf{Recall} & \textbf{F1} & \textbf{Precision} & \textbf{Recall} & \textbf{F1} \\ \midrule
\textbf{\textsc{ST-BOW}} & 78.00 & 78.00  & 78.00 & 57.00 & 56.00  & 57.00 & 67.50 & 67.00  & 67.50 \\
\textbf{\textsc{ST-All}} & 78.00 & 79.00  & 79.00 & 58.00  & 56.00 & 57.00 & 68.0  & 67.50 & 68.00 \\
\textbf{\textsc{ST-Disc}} & 80.00 & 79.00  & 80.00 & 58.00  & 56.00 & 57.00 & 68.00  & 67.50 & 68.00\\
\textbf{\textsc{LR-BOW}} & 69.00 & 65.00  & 67.00 & 53.00  & 57.00 & 55.00 & 61.00  & 61.00 & 61.00\\
\textbf{\textsc{LR-All}} & 79.00 & 70.00  & 74.00 & 52.00  & 64.00 & 58.00 & 65.50  & 67.00 & 66.00\\
\textbf{\textsc{LR-Disc}} &  75.00  & 84.00 & 80.00  & 60.00 & 45.00 & 52.00  & 67.50 & 64.50 & 66.00 \\ \midrule
\textbf{BERT} & 81.75 & 75.90  & 78.72 & 57.95 & 66.23  & 61.82 & 69.85 & 71.06  & 70.27 \\
\textbf{BERT+ENG} & 81.99 & 83.06  & \textbf{82.52} & 65.33  & 63.64 & \textbf{64.47} & 73.66  & 73.35 & \textbf{73.50} \\
\bottomrule
\end{tabular}
\caption{Results for the DesireDB task: identifying if a desire described in the document is fulfilled or not.}
\label{tab:desiredb_result}
\end{table*}


DesireDB~\cite{rahimtoroghi-etal-2017-modelling} is the task of predicting whether a desire expression is fulfilled or not, given its prior and posterior context. It requires aggregating information from multiple parts of the document. If a target desire is ``I want to be rich'', and the character's mental changed from ``sad'' to ``happy'' along the text, we can infer that their desire is likely to be fulfilled. 
%

We use the baseline systems described in \cite{rahimtoroghi-etal-2017-modelling}, based on SkipThought (ST) and Logistic Regression (LR), with manually  engineered lexical and discourse features.
%
We train a stronger baseline by encoding the prior and posterior context, as well as the desire expression, using BERT.
Then, we add an attention layer (Eq. \ref{eq:attn}) for the two contexts over the desire expression. 
The resulting three representations (the weighted prior and posterior representations, and the desire representation) are then concatenated. For ENG, we add an attention layer over  the nodes to form the ENG document representation.
We compare BERT and BERT+ENG document representations by feeding each of them into a two-layer FFN for classification, as described in Sec. \ref{sec:objectives} (Doc. Classification).

Table \ref{tab:desiredb_result} shows the result. The BERT baseline outperforms other baselines with a large gap, $4.27\%$ absolute increase in the averaged F1-score. Furthermore, BERT+ENG forms a better document summary for the target desire, which further increase another absolute $3.23\%$ on the avg. F1-score. These results illustrate that ENG can be used in various settings for modeling entity information.

\subsection{Qualitative Analysis}

\begin{figure*}[h]
\centering
\begin{subfigure}[t]{.7\linewidth}
\includegraphics[width=\linewidth]{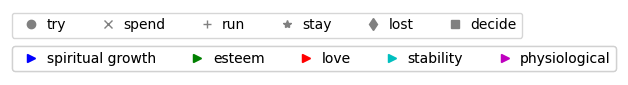}
\end{subfigure}
\vspace{-5px}
\newline
\begin{subfigure}[b]{.26\linewidth}
\includegraphics[width=\linewidth]{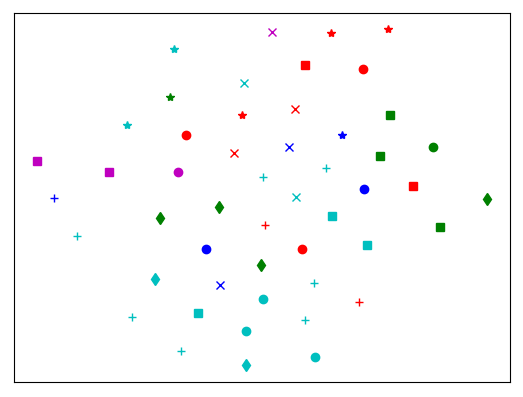}
\caption{ENG-CTX}\label{fig:tsne_eng}
\end{subfigure}
\begin{subfigure}[b]{.26\linewidth}
\includegraphics[width=\linewidth]{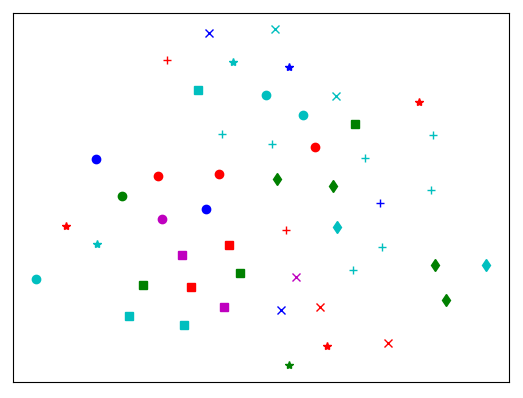}
\caption{W-CTX-STORY}\label{fig:tsne_nonctx_whole}
\end{subfigure}
\begin{subfigure}[b]{.26\linewidth}
\includegraphics[width=\linewidth]{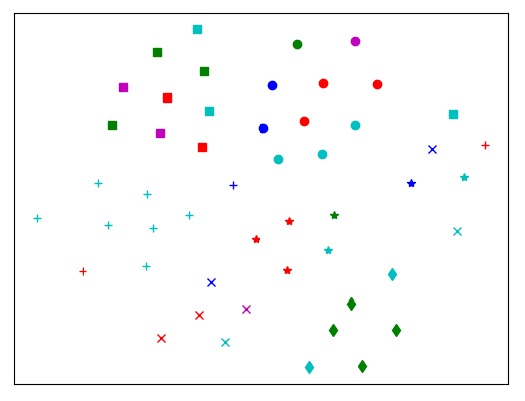}
\caption{W-CTX-SENT}\label{fig:tsne_nonctx}
\end{subfigure}
\caption{t-SNE visualization of embeddings based on ENG and RoBERTa.}
\label{fig:tsne}
\end{figure*}

We conduct a qualitative analysis by measuring and visualizing distances between event nodes corresponding to six verbs and their Maslow labels. We  project the node embeddings, based on different encoders, to a 2-D space using t-SNE~\cite{maaten2008visualizing}.  We use shapes to represent verbs and colors to represent labels. 
In Fig. \ref{fig:tsne_nonctx_whole} and \ref{fig:tsne_nonctx},  RoBERTa, pretrained on Whole-Word-Masking TAPT, was used.
Nodes are word-contextualized, receiving the whole story ({\small W-CTX-STORY}) or the target sentence ({\small W-CTX-SENT}) as context.
In these two cases, event nodes with the same verb (shape) tend to be closer. 
%
In Fig. \ref{fig:tsne_eng}, we use ENG as the encoder to generate graph-contextualized embeddings ({\small ENG-CTX}).
We observe that nodes with the same label (color) tend to be closer. In all cases, the embedding was trained using only the TAPT tasks, without task specific data. The ENG embedding is better at capturing entities' mental states, rather than verb information, as the graph structure is entity-driven.


\begin{figure}[ht]
\centering

\begin{subfigure}[b]{0.9\linewidth}
\includegraphics[width=\linewidth]{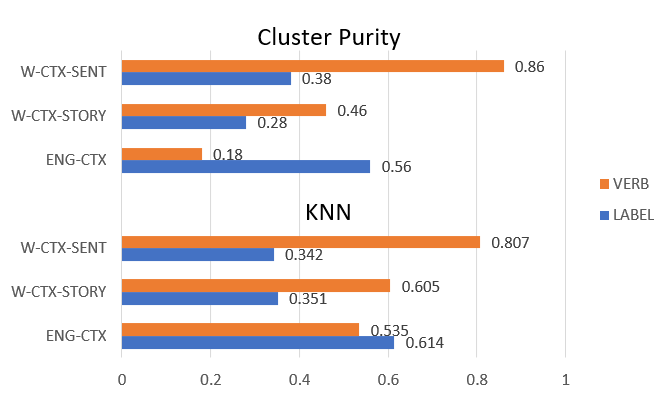}
\end{subfigure}


\caption{Cluster Purity and KNN Classification results for graph- and word-contextualized embeddings.}
\label{fig:cluster_bar}
\end{figure}

Figure \ref{fig:cluster_bar} makes this point quantitatively. We use 10-fold cross validation and report averaged results.
The proximity between verbs and between labels are measured in two ways: cluster purity and KNN classification. 
For the cluster purity~\cite{manning2008introduction}, we cluster the events using K-Means ($K=5$), and calculate the averaged cluster purity, defined as follows:
\begin{equation}
\frac{1}{N} \sum_{c \in C} \max_{d \in D} |c \cap d| ,
\end{equation}
where $C$ is the set of clusters and $D$ is either the set of labels or verbs.

For the graph contextualization, we can see that the labels have higher cluster purity than the verbs, while for the word contextualization, the verbs have higher cluster purity. This result aligns with our visualization.
The KNN classification uses the learned embedding as a distance function. The KNN classifier performs better when classifying labels using the graph-contextualized embeddings, while it performs better using word-contexualized embeddings when classifying verbs. These results demonstrate that ENG can better capture the states of entities.

\section{Conclusions}
We propose an ENG model that captures implicit information about the states of narrative entities using multi-relational graph contextualization.
We study three types of weakly-supervised TAPTs for ENG and their impact on the performance of downstream tasks, as well as symbolic inference capturing the interactions between predictions. Our empirical evaluation was done over two narrative analysis tasks. The results show that ENG can outperform other strong baselines, and the contribution of different types of TAPT is task-dependent. In the future, we want to connect different TAPT schemes and downstream tasks, and explore constrained representations.

\section{Acknowledgements}

We thank the reviewers for their efforts and insights.
This work was partially funded by the NSF and DARPA ASED program under contracts CNS-1814105 and 13000686.

\bibliography{anthology,custom}
\bibliographystyle{acl_natbib}

\end{document}